\documentclass[conference]{IEEEtran}
\usepackage{cite}

\ifCLASSINFOpdf
  \usepackage[pdftex]{graphicx}
  \graphicspath{{../pdf/}{../jpeg/}}
  \DeclareGraphicsExtensions{.pdf,.jpeg,.png}
  \usepackage{epstopdf}
\else
\fi
\usepackage{amsmath}
\usepackage{multirow}
\usepackage{algorithmic}
\usepackage{algorithm}

\usepackage{array}
\begin{document}
%
\title{Deep Template Matching for Offline Handwritten Chinese Character Recognition}

\author{\IEEEauthorblockN{Zhiyuan Li}
\IEEEauthorblockA{
lizhiyuan215@mails.ucas.edu.cn}
\and
\IEEEauthorblockN{Min Jin}
\and
\IEEEauthorblockN{Qi Wu}
\and
\IEEEauthorblockN{Huaxiang Lu}
}


%


\maketitle

\begin{abstract}
Just like its remarkable achievements in many computer vision tasks, the convolutional neural networks (CNN) provide an end-to-end solution in handwritten Chinese character recognition (HCCR) with great success. However, the process of learning discriminative features for image recognition is difficult in cases where little data is available. In this paper, we propose a novel method for learning siamese neural network which employ a special structure to predict the similarity between handwritten Chinese characters and template images. The optimization of siamese neural network can be treated as a simple binary classification problem. When the training process has been finished, the powerful discriminative features help us to generalize the predictive power not just to new data, but to entirely new classes that never appear in the training set. Experiments performed on the ICDAR-2013 offline HCCR datasets have shown that the proposed method has a very promising generalization ability to the new classes that never appear in the training set.
\end{abstract}


%
\IEEEpeerreviewmaketitle

\section{Introduction}
\label{intro}
Offline handwritten Chinese character recognition (HCCR) has been a important research realms since early works in 1980s\cite{kimura1987modified}. Due to the great diversity of handwriting style, confusion between similar characters and large number of character classes, offline HCCR is still a challenging problem. In the last few years, there has been a significant amount of work on improving HCCR performance. The typical recognition model for HCCR mainly focuses on three parts: preprocessing, feature extraction and classification. Although researchers have proposed many methods to improve the Chinese character recognition rate, the best traditional modified quadratic discriminant function (MQDF) based methods are still far from human performance. Benefits from the blooming growth of computational power, massive amounts of training data and better training technologies, deep convolutional neural networks (CNN)\cite{krizhevsky2012imagenet} have achieved significant improvement in many computer vision tasks. Nowadays, deep CNN-based approaches become the new novel technology for solving HCCR problems.

Most Chinese character recognition methods focus on a balanced dataset, which contains the frequently used 3755 characters in the GB2312-80 standard level-1 set and each character has hundreds of samples. All testing characters are shown at training time, which is known as a closed set recognition problem. However, a more complete set would contain about 7000 characters for modern Chinese texts. The number of characters is over 54000 for historical and scholarly collections, which corresponding to a open set recognition problem. To obtain a satisfying recognition performance, training samples for each character should be sufficient, especially for deep CNN-based methods. Therefore, current approaches can handle only a limited number of documents satisfactorily.          

In this paper, we propose a method to recognize Chinese characters as a template matching problem. The advantages of using template matching to recognize Chinese characters are: (1) In the current methods, the size of the model is proportional to the number of categories. Compared to predicting the probability that the character images falls into each category, our method has only one output unit which represent the similarity between templates and the input character images. Therefore, the size of our model is fixed no matter how many categories need to be classified. (2) We use sample pairs to train the template matching network, which will augment the training data naturally. For a $c$-class classification task, if there are n samples of each category in the training dataset, we can generate $nc^2$ sample pairs for template matching problem. The number of training samples increased to $c$ times of the original dataset. (3) Template matching can be used to recognize the Chinese characters which are not shown in the training set without any additional training process. This is well-known as zero-shot learning problem.

The rest of this paper is organized as follows. Section \ref{Related Works} reviews the related works about HCCR. Section \ref{Deep Siamese Network for Template Matching} introduces the details about the proposed method. Section \ref{Experiments}
reports the experimental results. The conclusions of this study and our future work are summarized in Section \ref{Conclusions}.

%
%
\section{Related Works}
\label{Related Works}
Due to the extraordinary achievement of deep learning in computer vision tasks\cite{krizhevsky2012imagenet,simonyan2015very,szegedy2015going,he2016deep} , the research for offline HCCR has been changed to convolutional neural networks (CNN). 
Multi-column deep neural networks (MCDNN)\cite{ciregan2012multi-column}\cite{ciresan2015multi-column} was the first reported successful use of CNN for offline HCCR. After that, a research team from Fujistu developed a CNN-based method and took the winner place in ICDAR-2013 competition\cite{yin2013icdar}. A voting format of alternately trained relaxation convolutional neural networks (ATR-CNN) was proposed by the same team in \cite{wu2014handwritten}. Zhong et al.\cite{zhong2015high} combined the traditional feature extraction methods with the inception architecture proposed in GoogLeNet\cite{szegedy2015going} and achieved very high accuracy for offline HCCR, which became the first model beyond human performance. In \cite{zhang2017online}, Zhang et al. added a adaptation layer into pre-trained CNN to adapt the new handwriting styles of particular writers, which sets new benchmarks for offline HCCR. 
Recently, some researchers have focused on the high computational cost and large storage requirement for CNN-based models.
 Xiao et al.\cite{xiao2017building} proposed a Global Supervised Low-rank Expansion and an Adaptive Drop-weight technique to solve
 the problems of speed and storage capacity. Li et al.\cite{li2018building} designed an efficient CNN architecture and implemented cascaded model in a single network, which achieved the state-of-the-art results for offline HCCR.

However, all current methods mentioned above consider each Chinese character as a single class and train a multi-class classifiers for HCCR (character-based classifier). These approaches require a great deal of labeled samples to optimize the models. And the learned model can not recognize new characters that are not shown in training datasets. In a real applications, the number of Chinese characters is very huge, and there will not be sufficient labeled data of the rarely-used characters for optimizing the models. Therefor, it is impossible to obtain a good performance using the character-based classifier in real-world applications.

\section{Deep Siamese Network for Template Matching}
\label{Deep Siamese Network for Template Matching}

\subsection{Siamese Network}

\begin{figure}[htb]
	\centering
	\includegraphics[width=18pc]{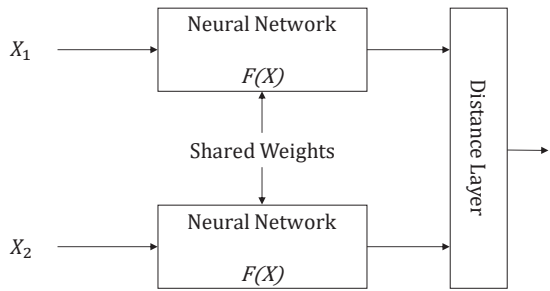}
	\caption{Siamese Neural Network.}
	\label{siamese}
\end{figure}

Siamese neural network is a class of neural network architectures that contain two subnetworks. Those subnetworks accept different inputs but share the same configuration with the same parameters and weights. An energy function is added at the top layer to compute some metric between the feature representation on each side. Parameter updating is mirrored across both subnetworks. A typical siamese neural network is shown in \figurename \ref{siamese}.

Siamese neural network was first introduced to solve signature verification by Bromley and Lecun in 1990s\cite{bromley1994signature} and became popular among tasks that involve finding similarity between two inputs. The network guarantees that two similar images should be mapped to adjacent location in feature space. In \cite{chopra2005learning}, Lecun et al. proposed a contrastive energy function which contained symmetric terms to decrease the energy of same pairs and increase the energy of different pairs.

\subsection{Template Matching}
\begin{figure}[htb]
	\centering
	\includegraphics[width=18pc]{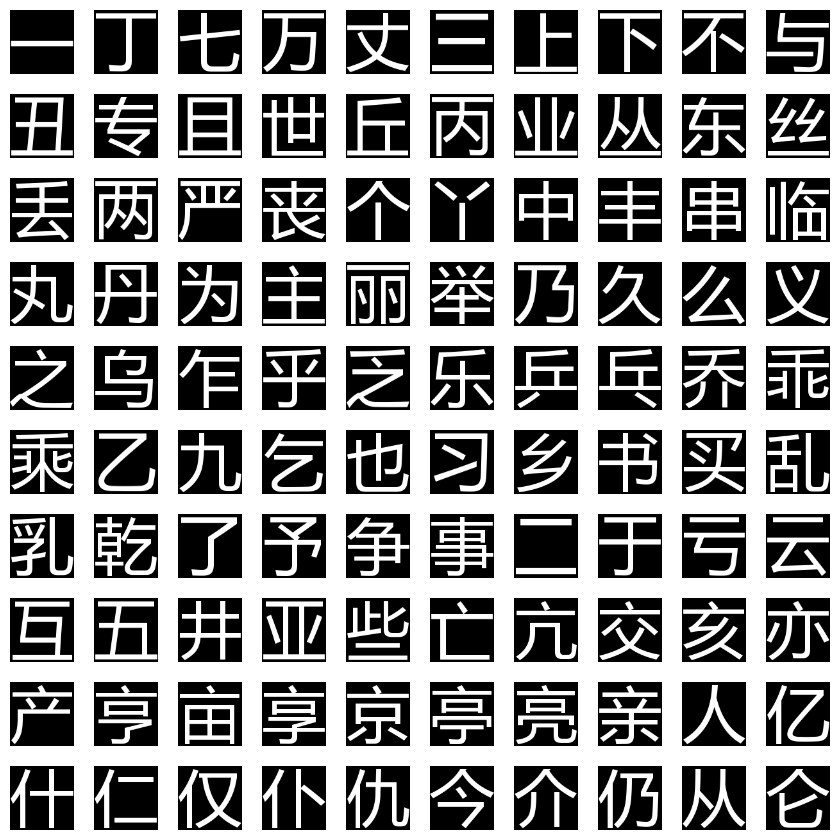}
	\caption{Examples of Chinese character templates.}
	\label{template}
\end{figure}

When we were learning Chinese, we always practiced writing by following the template character in the textbook. Once we learned how to write, we remembered this character forever. Inspired by this, we treat the HCCR task as a template matching problem. The template characters are generated by font of Microsoft YaHei (msyh.ttf). Some examples are shown in \figurename \ref{template}. 

Instead of using an energy function to compute some metric between the feature representation, we use the $L_1$ distance between the twin feature vectors $f_1$ and $f_2$ to predict the similarity between the templates and input character images. More precisely, the prediction is given by $p(I_x, I_c)=\sigma(w|f(I_x)-f(I_c)|+b)$, where $f(I)$ represents the feature vector for image $I$ extracted by the neural network, and $\sigma $ is the sigmoid activation function which maps the output onto the range [0,1]. Thus the template matching task can be treated as a binary classification problem and the binary cross entropy objective is a nature choice for training the network.

\begin{equation}
	\mathcal{L}(I_x,I_c,y)=y\log{p(I_x, I_c)}+(1-y)\log{p(I_x, I_c)}
\end{equation}

\subsection{Classification}
\begin{figure}[htb]
	\centering
	\includegraphics[width=18pc]{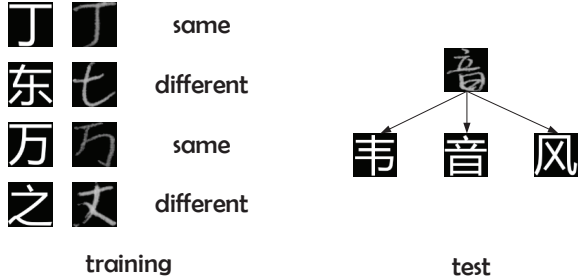}
	\caption{Template matching for offline HCCR. (1)Train a model to discriminate between template and handwritten character image. (2)Generalize to new characters based on the discriminative features learned by the siamese network.}
	\label{template_match}
\end{figure}

As shown in \figurename \ref{template_match}, when we have finished optimizing the siamese network as a binary classification task, we can use the discriminative capacity of the learned features for recognition problem. Suppose we are given a test image $I_x$, which we wish to classify into one of $C$ characters. We can generate those character template $\left\{I_c\right\}_{c=1}^{C}$, and query the network using $\{I_x, I_c\}$ as input pairs for a range of $c=1,2,\ldots,C$. Then predict the class to the maximum similarity.

\begin{equation}
	c^*=\arg\max_c{p(I, I_c)}
\end{equation}

Considering that the convolution neural network is usually accompanied by a huge amount of computation, we can extract the features for $C$ templates and stack them to form a matrix $F_C$. Once we have a new image $I_x$ to be classified, we can perform just one feedforward pass to extract the features $F_x$. Then we compute the similarity between $F_x$ and $F_C$ with negligible computation cost.   

\begin{figure*}[!t]
	\centering
	\includegraphics[width=40pc]{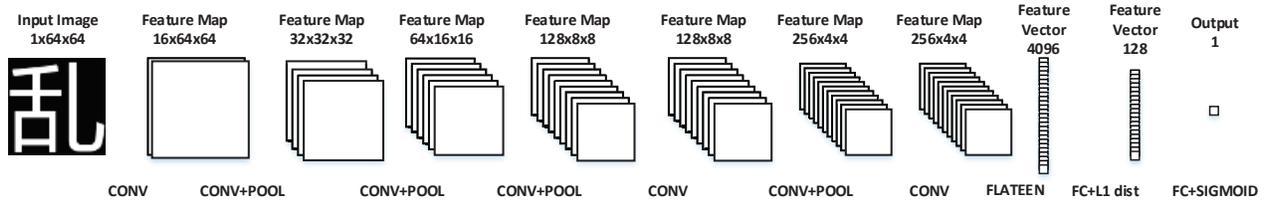}
	\caption{Baseline convolutional architecture for template matching problem.}
	\label{network}
\end{figure*}

\section{Experiments}
\label{Experiments}

\subsection{Datasets}
We use the offline CASIA-HWDB1.0 and CASIA-HWDB1.1 datasets for training our neural network, and evaluate the model performance on the ICDAR-2013 offline competition datasets. All datasets are collected by the Institute of Automation of the Chinese Academy of Sciences. The number of character classes is 3755(level-1 set of GB2312-80). Training dataset contains about 2.67 million samples contributed by 720 writers and test data contains about 0.22 million samples contributed by another 60 writers\cite{liu2011casia}.

\subsection{Architecture}
Limited by the hardware, we design a very compact architecture consisting of seven convolutional layers and one fully connected layers as base network. Each of the first three convolutional layers are followed by a max-pooling layer. Following this, two convolutional layers are followed by a max-pooling layer. Then another two convolutioinal layers are followed by a fully connected layer which contains 128 neurons. In our baseline network, batch normalization\cite{ioffe2015batch} is added for all convolutional layers to enable the network converge easily and minimize the risk of overfitting. The detailed architecture is shown in \figurename \ref{network}.

\subsection{Experimental Settings}
The datasets provide gray images with white background pixels. We process the gray images in three steps: firstly, we reverse the gray levels by $I^{\prime}=255-I$: background as 0 and foreground in a range of [1, 255]. Then we crop the redundant background area and preserve true boundary of the Chinese stroke. At last, the image is resized to $64\times64$ by contour center linear normalization with square root of sine of aspect\cite{liu2004handwritten}. Since batch normalization is added at all convolutional layers, we initialize the learning rate at 0.1, and then reduce it $\times0.1$ when the validation performance stops improving. We use stochastic gradient descent algorithm to train our model. The mini-batch is set to 256 with a momentum of 0.9. The regularization strategy we used is the weight decay with $L_2$ penalty. The multiplier for weight decay is set to $10^{-4}$ during the training process. We conduct all experiments on TensorFlow\cite{abadi2016tensorflow:} using a GTX1060 (3G) card.

\begin{algorithm}[h]
	\caption{Rules for generating sample pairs.}
	\label{sample}
	\begin{algorithmic}
		\STATE {initialize a sample list: $L=[]$}
		\FOR{$i=1$ to $c$}
		\FOR{$j=1$ to $c$}
		\IF{$i == j$} 
		\FOR{each $x \in D[j]$}
		\STATE L.append([$I_i, x, 1$])
		\ENDFOR
		\ELSE 
		\STATE random select $n$ samples in $D[j]$ as T
		\FOR{each $x \in T$}
		\STATE L.append([$I_i, x, 0$])
		\ENDFOR
		\ENDIF 	
		\ENDFOR
		\ENDFOR
		\STATE random shuffle the sample list $L$. 
	\end{algorithmic}
\end{algorithm}

\subsection{Performance on Open Set}
In these experiments, we split the Chinese characters into two parts: $\mathcal{C}_s$ and $\mathcal{C}_u$, which are two non-overlapping subset of all 3755 Chinese characters shown in CASIA-HWDB dataset. The datasets are also divided into two parts: $\mathcal{D}_s$ and $\mathcal{D}_u$. $\mathcal{D}_s$ contains all samples of $\mathcal{C}_s$ and $\mathcal{D}_u$ contains the rest images. The corresponding datasets are denoted as $\mathcal{D}_s^{train}$, $\mathcal{D}_u^{train}$, $\mathcal{D}_s^{test}$, $\mathcal{D}_u^{test}$ respectively. 

To train the siamese network, we sample positive and negative pairs from  $\mathcal{D}_s^{train}$. For a single Chinese character in $\mathcal{C}_s$, we use all samples labeled as this character to generate positive pairs. For the remaining Chinese characters, we randomly select $n$ samples for each characters to generate negative pairs. The specific sampling method is listed in Algorithm \ref{sample}. 
We use 3 character sizes of $\mathcal{C}_s$ and 3 sampling ratios, yielding 9 different dataset. To monitor performance during training, we use the recognition accuracy of unseen characters on $\mathcal{D}_u^{train}$ as validation criteria to determine when to reduce the learning rate or stop training. After the training process is finished, we evaluate the recognition rate on  $\mathcal{D}_s^{test}$, $\mathcal{D}_u^{test}$, $\mathcal{D}^{test}$ respectively.
The final classification results for each of the 9 training sets are listed in the table below (\tablename{ \ref{experiment}}). These results shows that our proposed method have a promising generalization ability to new Chinese characters.

\begin{table}[htb]
	\renewcommand{\arraystretch}{1.3}
	\caption{Performance on CASIA competitionDB
		($\mathcal{A}|\mathcal{B}$ means recognize dataset $\mathcal{A}$ on character set $\mathcal{B}$).}
	\label{experiment}
	\centering
	\begin{tabular}{|c|c|c|c|c|c|c|}
		\hline
		\multirow{2}{*}{\bfseries c} & \multirow{2}{*}{ \bfseries n} & \multicolumn{5}{c}{ \bfseries accuracy(\%)} \\
		\cline{3-7}
		\multirow{2}{*}{} & \multirow{2}{*}{} & $\mathcal{D}_s|\mathcal{C}_s $ & $\mathcal{D}_s|\mathcal{C}$ &  $\mathcal{D}_u|\mathcal{C}_u$ & $\mathcal{D}_u|\mathcal{C}$ & $\mathcal{D}|\mathcal{C}$ \\
		\hline
		\multirow{3}{*}{500} & 5 & 89.88 & 77.67 & 44.90 & 43.97 & 48.44 \\
		\cline{2-7}
		\multirow{3}{*}{} & 10 & 91.60 & 81.41 & 46.73 & 45.70 & 50.44\\
		\cline{2-7}
		\multirow{3}{*}{} & 20 & 91.90 & 81.57 & 48.56 & 47.61 & 52.11\\
		\hline
		\multirow{3}{*}{1000} & 5 & 90.31 & 82.31 & 62.00 & 60.10 & 66.01\\
		\cline{2-7}
		\multirow{3}{*}{} & 10 & 91.06 & 84.59 & 62.87 & 60.64 & 67.01\\
		\cline{2-7}
		\multirow{3}{*}{} & 20 & 90.67 & 85.04 & 61.85 & 59.24 & 66.10\\
		\hline
		\multirow{3}{*}{2000} & 5 & 88.07 & 84.81 & 75.95 & 71.01 & 78.36\\
		\cline{2-7}
		\multirow{3}{*}{} & 10 & 88.56 & 86.02 & 77.37 & 71.67 & 79.31\\
		\cline{2-7}
		\multirow{3}{*}{} & 20 & 89.22 & 86.51 & 77.85 & 72.72 & 80.06\\
		\hline
	\end{tabular}	
	
\end{table}

\subsection{Performance on Close Set}
In this section, we conduct experiments on all 3755 Chinese characters. For fair comparison, we train the siamese network as a template-matching classifier, then we replace the output layer with a fully connected layer which contains 3755 units to perform classification. The second network (character-based classifier) is trained as a 3755-way classification task using the traditional cross entropy loss function. Table \ref{comparision} provides the recognition accuracy of different models using ICDAR-2013 offline HCCR competition datasets.
 
\begin{table}[htb]
	\renewcommand{\arraystretch}{1.3}
	\caption{Different methods for ICDAR-2013 offline HCCR competition.}
	\label{comparision}
	\centering
	\begin{tabular}{c c}
		\hline
		Method & Accuracy \\
		\hline
		DFE+DLQDF & 92.72 \\
		\hline
		CNN-Fujitsu & 94.77 \\
		\hline
		MCDNN & 95.79 \\
		\hline
		ATR-CNN & 96.06 \\
		\hline
		Human & 96.13 \\
		\hline 
		HCCR-Gabor-GoogLeNet & 96.35 \\
		\hline
		DirectMap+CNN & 96.95 \\
		\hline
		HCCR-CNN9 & 97.09 \\
		\hline\hline
		Character-based Classifier & 95.51 \\
		\hline
		Template-Matching Classifier & 92.31 \\
		\hline	
	\end{tabular}	
\end{table}

\begin{figure}[htb]
	\centering
	\includegraphics[height=23pc]{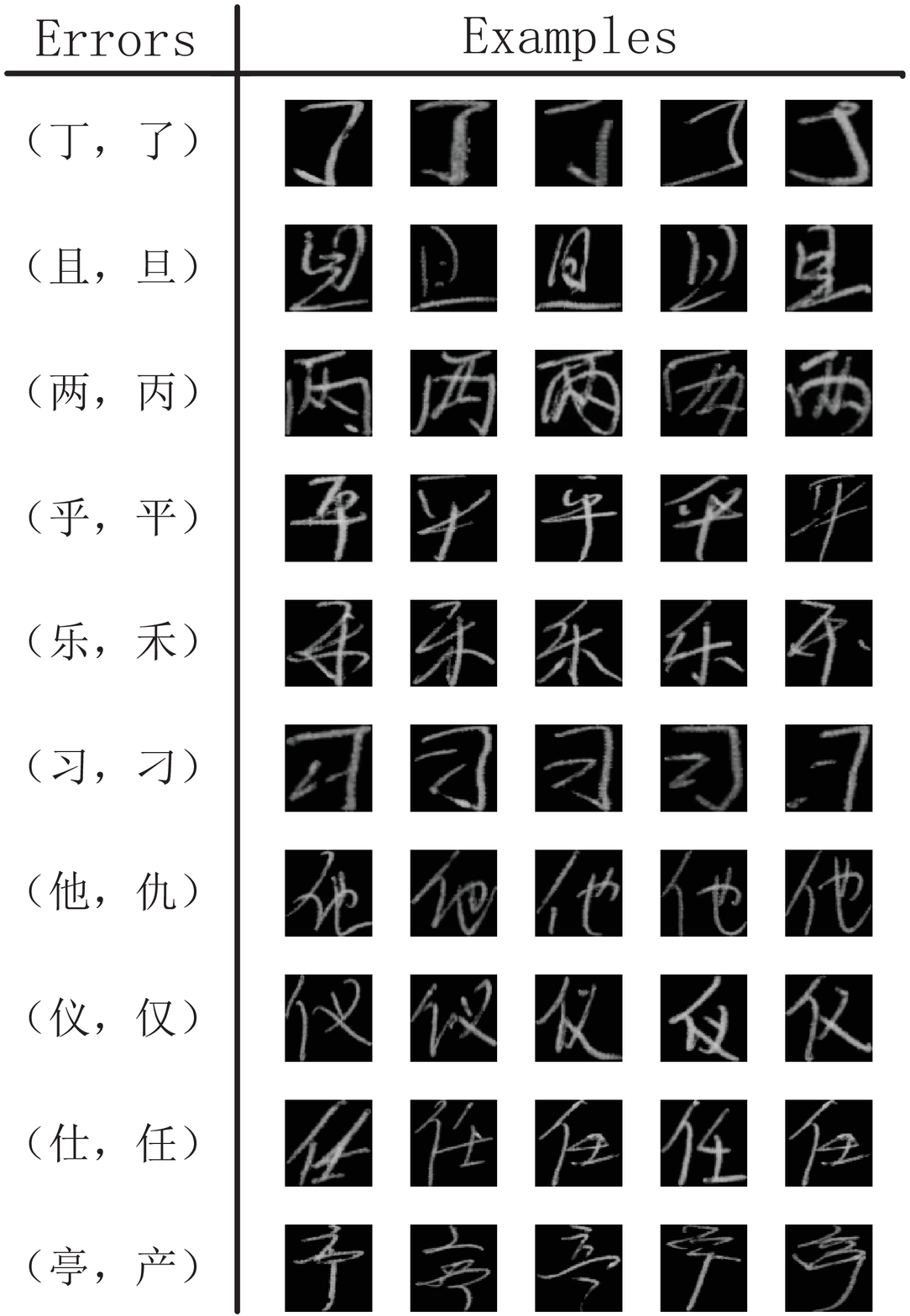}
	\caption{Top-10 failure examples of character recognition on the ICDAR-2013 offline competition datasets. The left column is (ground truth, prediction).}
	\label{error}
\end{figure}

Our proposed template-matching-based classifier provides lower accuracy than the current methods because template-matching-based neural network learns the similarity metric on the characters, which is very difficult for handwritten Chinese characters, especially on the similar characters. \figurename{\ref{error}} shows the top-10 false predictions of our template-matching-based classifiers. As can be seen, the recognized characters are very similar to the ground truth.

\section{Conclusions}
\label{Conclusions}
In this paper, we present a new technology for handwritten Chinese character recognition by learning deep siamese convolutional neural networks for template matching. The character templates are machine-printed images which uses Microsoft YaHei font. We evaluate our template-matching-based recognition on CASIA-HWDB dataset. The results shows that our proposed method can recognize characters which are not shown in the training set. To the best of our knowledge, no research has focused on the work of handwritten Chinese character recognition for "open set". In our future work, we will focus on better understanding of the error cases, further improving the model and reducing the performance gap with state of the art methods.


\section*{Acknowledgment}

This paper is supported by Strategic Pilot Special Science and Technology Project of Chinese Academy of Sciences: XDA18000000.




\begin{thebibliography}{1}
  
\bibitem{kimura1987modified}
F.~Kimura, K.~Takashina, S.~Tsuruoka, and Y.~Miyake, ``Modified quadratic
discriminant functions and the application to chinese character
recognition,'' \emph{IEEE Transactions on Pattern Analysis and Machine
	Intelligence}, vol.~9, no.~1, pp. 149--153, 1987.

\bibitem{liu2013online}
C.~Liu, F.~Yin, D.~Wang, and Q.~Wang, ``Online and offline handwritten chinese
character recognition: Benchmarking on new databases,'' \emph{Pattern
	Recognition}, vol.~46, no.~1, pp. 155--162, 2013.

\bibitem{yong2002chinese}
G.~Yong, H.~Qiang, and F.~Zhidan, ``Chinese character recognition: History,
status, and prospects,'' \emph{n: Proceedings of the 2002 IEEE International
	Conference on Acoustics, Speech, and Signal Processing. Orlando, FL, USA:
	IEEE}, 2002.

\bibitem{liu2004handwritten}
C.~Liu, K.~Nakashima, H.~Sako, and H.~Fujisawa, ``Handwritten digit
recognition: investigation of normalization and feature extraction
techniques,'' \emph{Pattern Recognition}, vol.~37, no.~2, pp. 265--279, 2004.

\bibitem{liu2007normalization-cooperated}
C.~Liu, ``Normalization-cooperated gradient feature extraction for handwritten
character recognition,'' \emph{IEEE Transactions on Pattern Analysis and
	Machine Intelligence}, vol.~29, no.~8, pp. 1465--1469, 2007.

\bibitem{mangasarian2002data}
O.~L. Mangasarian and D.~R. Musicant, ``Data discrimination via nonlinear
generalized support vector machines,'' \emph{Comple-mentarity: Applications,
	Algorithms and Extensions. US:Springer}, pp. 233--251, 2001.

\bibitem{liu2004discri}
C.~Liu, Sako.~H, and Fujisawa.~H, ``Discriminative learning quadratic
discriminant function for handwriting recogni-tion,'' \emph{IEEE Transactions
	on Neural Networks}, vol.~15, no.~2, pp. 430--444, 2004.

\bibitem{krizhevsky2012imagenet}
A.~Krizhevsky, I.~Sutskever, and G.~E. Hinton, ``Imagenet classification with
deep convolutional neural networks,'' pp. 1097--1105, 2012.

\bibitem{simonyan2015very}
K.~Simonyan and A.~Zisserman, ``Very deep convolutional networks for
large-scale image recognition,'' \emph{international conference on learning
	representations}, 2015.

\bibitem{szegedy2015going}
C.~Szegedy, W.~Liu, Y.~Jia, P.~Sermanet, S.~E. Reed, D.~Anguelov, D.~Erhan,
V.~Vanhoucke, and A.~Rabinovich, ``Going deeper with convolutions,''
\emph{computer vision and pattern recognition}, pp. 1--9, 2015.

\bibitem{he2016deep}
K.~He, X.~Zhang, S.~Ren, and J.~Sun, ``Deep residual learning for image
recognition,'' \emph{computer vision and pattern recognition}, pp. 770--778,
2016.

\bibitem{ioffe2015batch}
S.~Ioffe and C.~Szegedy, ``Batch normalization: Accelerating deep network
training by reducing internal covariate shift,'' \emph{international
	conference on machine learning}, pp. 448--456, 2015.

\bibitem{ciregan2012multi-column}
D.~Ciregan, U.~Meier, and J.~Schmidhuber, ``Multi-column deep neural networks
for image classification,'' \emph{computer vision and pattern recognition},
pp. 3642--3649, 2012.

\bibitem{ciresan2015multi-column}
D.~C. Ciresan and U.~Meier, ``Multi-column deep neural networks for offline
handwritten chinese character classification,'' \emph{international symposium
	on neural networks}, pp. 1--6, 2015.

\bibitem{yin2013icdar}
F.~Yin, Q.~Wang, X.~Zhang, and C.~Liu, ``Icdar 2013 chinese handwriting
recognition competition,'' \emph{Proc. Int’l Conf. Document Analysis and
	Recognition (ICDAR)}, pp. 1095--1101, 2013.

\bibitem{wu2014handwritten}
C.~Wu, W.~Fan, Y.~He, J.~Sun, and S.~Naoi, ``Handwritten character recognition
by alternately trained relaxation convolutional neural network,'' pp.
291--296, 2014.

\bibitem{zhong2015high}
Z.~Zhong, L.~Jin, and Z.~Xie, ``High performance offline handwritten chinese
character recognition using googlenet and directional feature maps,''
\emph{international conference on document analysis and recognition}, pp.
846--850, 2015.

\bibitem{zhang2017online}
X.~Zhang, Y.~Bengio, and C.~Liu, ``Online and offline handwritten chinese
character recognition: A comprehensive study and new benchmark,''
\emph{Pattern Recognition}, vol.~61, pp. 348--360, 2017.

\bibitem{xiao2017building}
X.~Xiao, L.~Jin, Y.~Yang, W.~Yang, J.~Sun, and T.~Chang, ``Building fast and
compact convolutional neural networks for offline handwritten chinese
character recognition,'' \emph{Pattern Recognition}, vol.~72, 2017.

\bibitem{li2018building}
Z.~Li, N.~Teng, M.~Jin, and H.~Lu, ``Building efficient CNN architecture for offline handwritten Chinese character recognition,'' \emph{International Journal on Document Analysis and Recognition}, vol.~21, 2018.

\bibitem{liu2011casia}
C.~Liu, F.~Yin, D.~Wang, and Q.~Wang, ``Casia online and offline chinese
handwriting databases,'' pp. 37--41, 2011.

\bibitem{bromley1994signature}
J.~Bromley, I.~Guyon, Y.~Lecun, E.~Sackinger, and R Shah, ``Signature Verification using a "Siamese" Time Delay Neural Network,'' \emph{neural information processing systems}, pp. 737--744, 1994.

\bibitem{chopra2005learning}
S.~Chopra, R.~Hadsell, and Y.~Lecun, ``Learning a similarity metric discriminatively, with application to face verification,'' \emph{computer vision and pattern recognition}, pp. 539--546, 2005.

\bibitem{abadi2016tensorflow:}
M.~Abadi, P.~Barham, J.~Chen, Z.~Chen, A.~Davis, J.~Dean, M.~Devin,
S.~Ghemawat, G.~Irving, M.~Isard \emph{et~al.}, ``Tensorflow: a system for
large-scale machine learning,'' \emph{operating systems design and
	implementation}, pp. 265--283, 2016.

\end{thebibliography}
%

\end{document}